\pdfoutput=1
\documentclass[11pt]{article}

\usepackage{acl}

\usepackage{times}
\usepackage{latexsym}
\usepackage{amsmath}

\usepackage{amssymb}
\usepackage{mathabx}
\usepackage{graphicx}
\usepackage{multirow}
\usepackage{multicol}
\usepackage{stfloats}
\usepackage{float}
\usepackage{caption}
\usepackage[T1]{fontenc}
\usepackage{xcolor}
\usepackage{url}
\usepackage{makecell}
\usepackage{pifont}
\usepackage{booktabs}
\usepackage{balance}

\usepackage{longtable}
\usepackage{caption}
\usepackage{multirow}
\usepackage{booktabs}

\usepackage[utf8]{inputenc}

\usepackage{microtype}

\title{Exploring Automated Distractor Generation for \\ Math Multiple-choice Questions via Large Language Models}
\author{
\bf Wanyong Feng$^{1*}$, Jaewook Lee$^{1*}$, Hunter McNichols$^{1*}$, Alexander Scarlatos$^{1*}$, \\
\bf Digory Smith$^2$, Simon Woodhead$^2$, Nancy Otero Ornelas$^3$, Andrew Lan$^1$ \\
  University of Massachusetts Amherst$^1$, Eedi$^2$,  Stanford University, Kitco Design$^3$\\
  \texttt{\{wanyongfeng,ajscarlatos,jaewooklee,wmcnichols,andrewlan\}@umass.edu} \\
  \texttt{\{digory.smith,simon.woodhead\}@eedi.co.uk} \\
  \texttt{nancy@kitco.design}
  }

\begin{document}
\maketitle
\def\thefootnote{*}\footnotetext{These authors contributed equally to this work.}
\begin{abstract}
Multiple-choice questions (MCQs) are ubiquitous in almost all levels of education since they are easy to administer, grade, and are a reliable format in assessments and practices. One of the most important aspects of MCQs is the distractors, i.e., incorrect options that are designed to target common errors or misconceptions among real students. To date, the task of crafting high-quality distractors largely remains a labor and time-intensive process for teachers and learning content designers, which has limited scalability. In this work, we study the task of automated distractor generation in the domain of math MCQs and explore a wide variety of large language model (LLM)-based approaches, from in-context learning to fine-tuning. We conduct extensive experiments using a real-world math MCQ dataset and find that although LLMs can generate some mathematically valid distractors, they are less adept at anticipating common errors or misconceptions among real students.
\end{abstract}

\section{Introduction}
\begin{figure*}[t]
\centering
\includegraphics[width=1\linewidth,trim=4 4 4 4,clip]{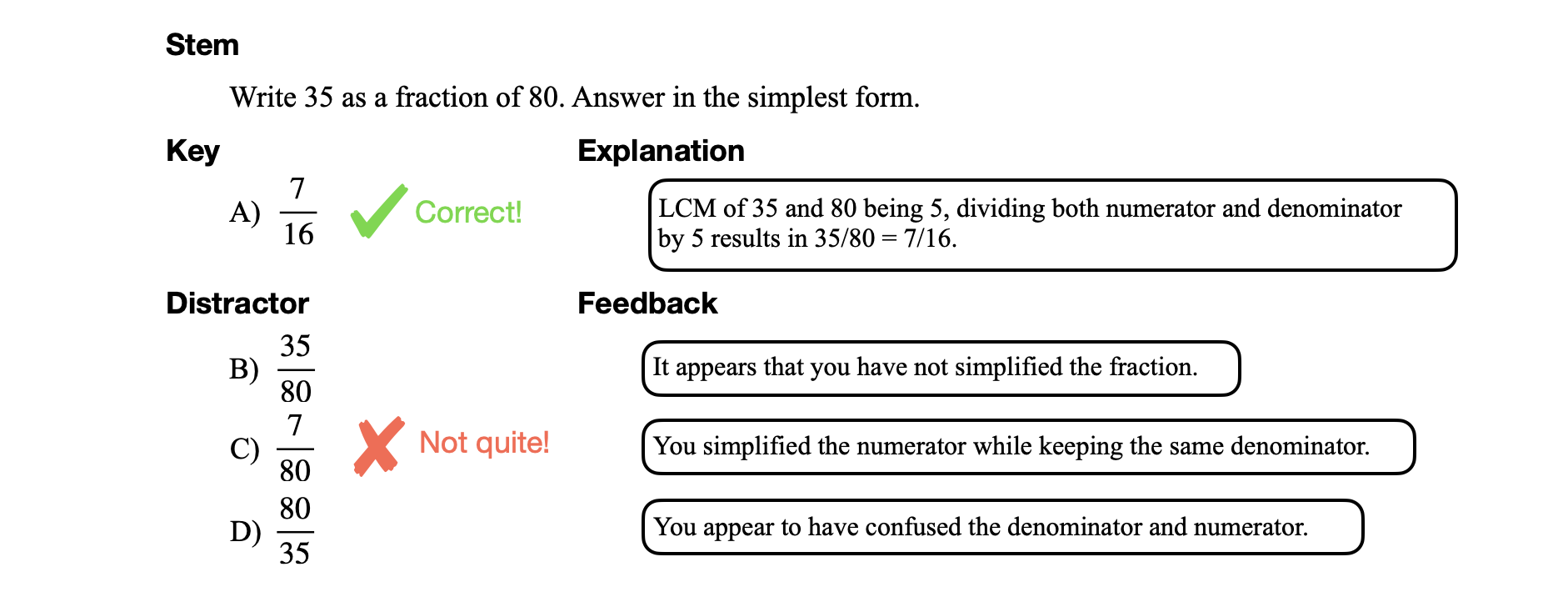}
\caption{Different parts of math MCQs and the terminology we use, illustrated with an example.}
\label{fig:terminology}
\vspace{-0.35cm}
\end{figure*}

Multiple-choice questions (MCQs) are widely used to evaluate student knowledge because they enable quick and accurate administration and grading. MCQs are reliable because they are designed to measure specific learning objectives consistently \cite{Nitko:96, Airasian:01, Kubiszyn:16}. MCQs are constructed in a specific format. See Figure~\ref{fig:terminology} for an example. The \textit{stem} refers to the statement on the problem setup and context, followed by a question that needs to be answered. Among the options, the correct one can be referred to as the \textit{key}, while incorrect ones can be referred to as \textit{distractors}. As the name implies, distractors in MCQs are typically formulated to align with the common errors students would make or misconceptions students would exhibit. These distractors are chosen because students either i) lack the necessary knowledge of the skills tested in the MCQ to accurately identify the key as the correct answer, or ii) hold misconceptions that result in selecting a specific distractor as the correct answer.
While MCQs offer many advantages for student knowledge evaluation, manually crafting high-quality MCQs is a demanding and labor-intensive process~\cite{kelly2013traditional}. Specifically, high-quality distractors should be plausible enough to mislead students and not so evidently incorrect to be identified easily.

Prior work on automatic distractor generation primarily focuses on language learning and reading comprehension tasks, where distractors are used to assess students' comprehension of a given text or article. Early works use a ranking approach based on semantic similarity and word collocation information or a pre-defined ontology to produce distractors \cite{susanti2018automatic, stasaski2017multiple,alsubait2014generating}. More recent works use encoder-decoder models with attention mechanisms for distractor generation, resulting in longer and higher-quality distractors \cite{qiu2020automatic, shuai2023qdg, xie2021diverse, gao2019generating}. Additionally, several recent works use pre-trained large language models (LLMs) such as \texttt{BERT} and \texttt{T5} for distractor generation in the context of Swedish reading and Cloze test \cite{kalpakchi2021bert, chiang2022cdgp, rodriguez2022end}. Other works prompt LLMs such as \texttt{ChatGPT} and \texttt{GPT-4} to generate distractors, either by providing detailed instructions or in-context examples in their prompts, for computer science course quiz questions and questions testing language mastery or factual knowledge \cite{tran2023generating,bitew2023distractor}.

However, there is limited work on automatic distractor generation for math MCQs. This problem is more challenging than generating distractors for reading comprehension tasks because plausible distractors are not necessarily contained or can be inferred from the passage. A model for math MCQ distractor generation should have some math problem-solving capability and more importantly, an understanding of the common errors or misconceptions among real students. Existing works either use constraint logic programming \cite{tomas2013automatic} or manually constructed rules \cite{prakash2023q} to generate distractors. However, these works only applies to math MCQs generated by templates. The work in \cite{dave2021math} explores generating distractors using a neural network. However, their approach is training a math problem solver model and treating the incorrect outputs as distractors, which cannot capture common errors or misconceptions among real students. 
\subsection{Contributions}
In this work, we investigate the task of automatically generating plausible distractors for math MCQs using LLMs. Our contributions include:
\begin{itemize}
    \item We explore a variety of approaches to this task, including in-context learning, fine-tuning, and chain-of-thought prompting, together with rule- and sampling-based baselines. \footnote{Our code is publicly available at \url{https://github.com/umass-ml4ed/prompt_distractor_generation_NAACL}}
    \item We conduct extensive quantitative and qualitative experiments on a real-world dataset of math MCQs. We find that the most effective approach is in-context learning, where we select a few example MCQs as input to the LLM, which can serve as a baseline for future work. 
    \item We conduct a human evaluation and find that although the LLM-generated distractors are close to the human-authored ones in terms of mathematical validity, they do not necessarily reflect common errors or misconceptions among real students. 
\end{itemize}  

\section{Task and Approaches}
\label{sec:proposed_task}
In this section, we first formally define relevant mathematical notation in MCQs and the automated distractor generation task. We then detail the LLM-based approaches and baselines that we explore. 

\subsection{Task Definition}
We define an MCQ $Q$ as a set of textual components, i.e., $Q = \{s, k, e_k, D, F\}$.\footnote{In this paper, we do not consider MCQs that contain diagrams or images; extending our work to multi-modal MCQ content is left for future work.} Each MCQ contains a stem $s$, a key $k$, an (optional) explanation of the key $e_k$, and a set of distractors $D$; each of which has an (optional) corresponding feedback message $f_i$ which is shown to a student upon selecting a distractor $d_i \in D$. All of these components are sequences of words and math symbols (e.g., $s=\{w_1, \dots, w_L\}$ where $L$ is the length of the sequence $s$). 
Similar to \cite{qiu2020automatic}, we formulate the task of distractor generation as learning a function $g^\text{dis}$ that outputs a set of distractors $\hat{D}$ for an MCQ given the question stem and (optionally) key and its explanation, i.e.,
\begin{align} \label{eq:dgen}
g^\text{dis}(s, k, e_k) \rightarrow \hat{D}.
\end{align} 
Our goal is to generate distractors that students with insufficient knowledge on skills required for the MCQ or specific misconceptions will select. This way, the MCQ can better distinguish between students that master all the required skills and those who do not. Below, we detail various LLM-based distractor generation approaches and several baselines that we explore. 

We note that in this work, we study the problem of generating a set of distractors $\hat{D}$ given a single question stem . This setting is different from a possible alternative setting where we generate distractors one-by-one, each corresponding to a common error or misconception among real students. The latter is applicable to the related problem of \textit{feedback generation} \cite{prihar2023comparing}, which investigates the task of generating a feedback message $f_i$ for a distractor $d_i$. Providing feedback messages to students who select distractors can help them identify their errors or misconceptions and guide them towards the correct answer, which may expedite their learning process. In this work, we only treat the feedback message as an additional reasoning pathway to help LLMs generate plausible distractors and do not study the quality of feedback messages, which we leave for future work. 

\begin{figure*}[t]
\centering
\includegraphics[width=1\linewidth,trim=4 4 4 4,clip]{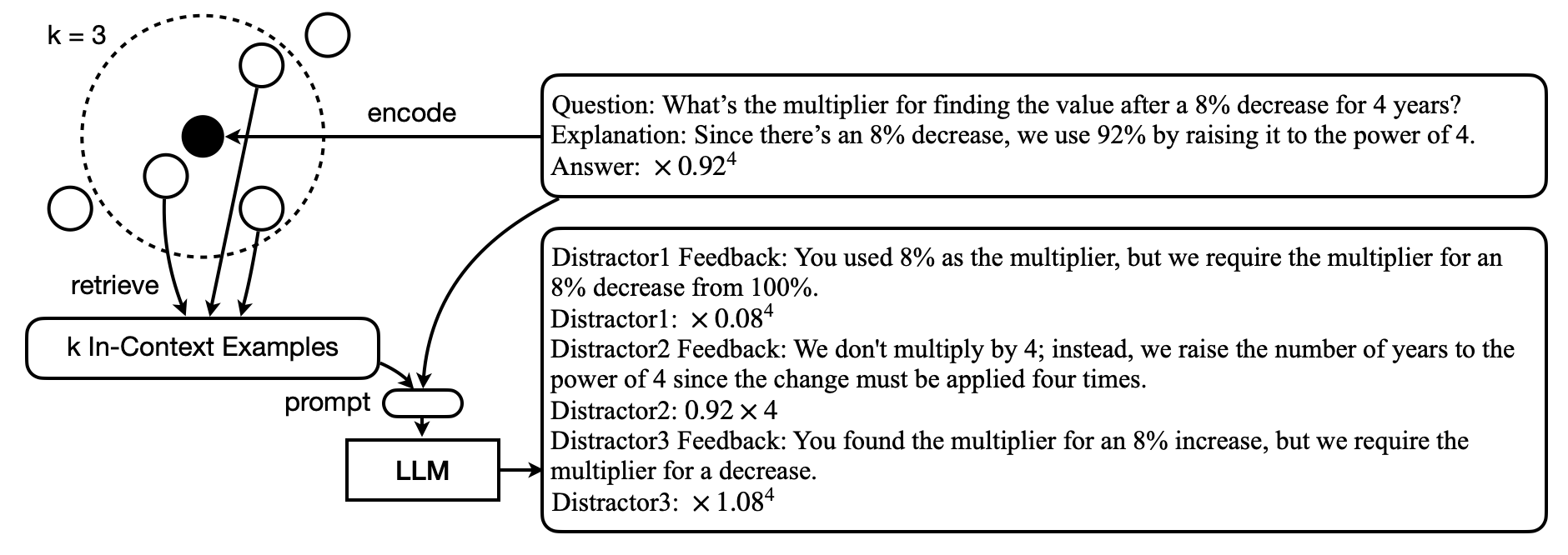}
\caption{Overview of the kNN approach illustrated with a math MCQ on ``compound percentage decrease''.}
\label{fig:knn_example}
\vspace{-0.35cm}
\end{figure*}

\subsection{Approaches}
The first approach is in-context learning or few-shot prompting, i.e., the LLM is expected to generate desired outputs for a new task by learning from the given examples  \cite{brown2020language}. To select examples, we select the k-nearest neighbor (\textbf{kNN}) MCQs from a real-world math MCQ dataset, which we detail in Section~\ref{sec:dataset}, to the target MCQ. After conducting tests with various values of $k$, we find that this approach achieves the best distractor generation performance when $k=3$. To determine similarity, we calculate the \textit{cosine similarity} between vectorized textual encodings of MCQs. Specifically, we use the pre-trained SBERT encoder \texttt{MPNet} \cite{reimers2019sentencebert} to calculate the textual encoding of the question stem and (optionally) key and its explanation. Figure~\ref{fig:knn_example} provides a visual representation of this approach. The intuition for this approach is that MCQs with similar question stems may have distractors that correspond to similar student errors or misconceptions that are feasible to the target MCQ, which may help the LLM to generate plausible distractors. Even though textual similarity may not be an appropriate representation for mathematical errors, these in-context examples should at least inform the LLM on distractor formatting \cite{chen2023exploring,lyu-etal-2023-z}. We use \texttt{ChatGPT} in this approach for its proficiency in understanding tasks and delivering strong performance when provided with in-context examples.

The second approach is chain-of-thought prompting (\textbf{CoT}) \cite{wei2022cot}. We provide the LLM with the question stem and (optionally) key and its explanation and detailed guidelines on distractor generation as input and ask it to first generate potential erroneous steps a student may take, followed by an incorrect answer as the distractor. This approach operates in a zero-shot manner and requires no access to any real MCQ data. Therefore, the performance depends solely on the LLM's ability in mathematical reasoning and anticipating common errors or misconceptions among real students. Given the demanding nature of this approach, we use a strong base LLM \texttt{GPT-4} \cite{openai2023gpt4}.

The third approach is LLM fine-tuning (\textbf{FT}) to help pre-trained LLMs to adapt to the distractor generation task. We use the real-world math MCQ dataset to fine-tune the LLM in the format of Eq.~\ref{eq:dgen}, i.e., outputting all distractors given the question stem and (optionally) key and its explanation as input. We use \texttt{ChatGPT} (gpt-3.5-turbo-1106) \cite{chatgpt}, the largest base LLM that can be fine-tuned, in this approach.

% This approach requires a strong base LLM so we use \texttt{GPT-4} \cite{bubeck2023sparks}.

The fourth approach is a rule-based (\textbf{RB}) baseline, which can be used to generate different versions of the same MCQ with different numerical values. We emphasize that in many real-world educational platforms, content creators do not use rules to design distractors. In practice, not a lot of MCQs are created from templates and only differ by numerical values or named entities in their question stems. Therefore, we \emph{approximately} follow the baseline approach in \cite{dave2021math} and manually construct $444$ distinct error explanations, such as ``confuses factor and multiples'' for question-distractor pairs that correspond to common errors or misconceptions among real students. This process is extremely time-consuming and requires significant manual effort. We then provide the LLM with the question stem and (optionally) key and its explanation and a pool of error explanations that are feasible under the MCQ's topic (i.e., fractions, rounding, etc.), and ask LLM to select 3 relevant ones and generate the corresponding distractors. We use \texttt{GPT-4} in this approach for the same reason as CoT. 

The fifth approach is an improved version of the sampling-based (\textbf{SB}) baseline in \cite{dave2021math}. This approach fine-tunes a base LLM on MCQ answering, i.e., outputting the key given the question stem as input. Then, we randomly sample up to 20 output answers from the trained LLM given a question stem as input and choose 3 distinct incorrect ones as distractors. This approach implicitly assumes that LLMs make similar errors as real students. We use \texttt{ChatGPT} in this approach for the same reason as FT.

\section{Experiments}
In this section, we detail the specifics of our dataset, the evaluation metrics, the experimental setup, and report results from a series of quantitative, qualitative experiments, and human evaluation. 

\subsection{Dataset}
\label{sec:dataset}
Our dataset consists of 1.4K MCQs from Eedi's content repository\footnote{\url{https://eedi.com/home}}, and all MCQs are written in English. Each question has 1 key and 3 distractors designed according to common errors or misconceptions among real students. The questions are sourced from the broad mathematical topic titled ``Number'' with subtopics including ``Basic Arithmetic'', ``Fractions'', and ``Rounding and Estimating''. The questions are primarily targeted towards students aged between 10 to 13. Each MCQ also has some additional metadata, e.g., the ``topic'' on 3 different granularity levels and the option selection distribution, i.e., the proportion of students who selected each option. The option selection distribution is computed on an average of 4000 student responses, with more than 900 student responses available in over 75\% of the MCQs. We divide the dataset into two subsets, namely a training set and a test set, using an $80:20$ ratio. We use the training set to select MCQs as in-context examples or fine-tune LLMs and the test set for evaluation. 

\subsection{Evaluation Metrics}
\label{sec:eval_metrics}
Our main evaluation metric is a set of \textit{alignment-based} metrics, which quantifies the extent to which the LLM-generated distractors align with the human-authored ones. We denote the LLM-generated distractors as $\hat{D}$ where $|\hat{D}|=N$.  We utilize 3 measures for this evaluation, two binary and one continuous. 
%Prior to computing these metrics, the predicted and ground truth sets must be index-aligned. 
The binary metrics are \textbf{Exact} match $h_e$, i.e., whether all LLM-generated distractors match human-authored ones\footnote{We use the exact string match criterion to align LLM-generated distractors with human-authored ones.},  and \textbf{Partial} match $h_p$, i.e., whether at least one LLM-generated distractor matches human-authored ones. These measures are formally defined as
\begin{align*}
    h_e(D,\hat{D}) = \begin{cases} 
        1 & \forall{\hat{d_i}} \in{\hat{D}} : \hat{d_i} = d_i \\
        0 & \text{otherwise}.
    \end{cases}
\end{align*}
and
\begin{align*}
    h_p(D,\hat{D}) = \begin{cases} 
      1 & \exists{\hat{d_i}} \in{\hat{D}} : \hat{d_i} = d_i \\
      0 & \text{otherwise}
    \end{cases}
\end{align*}

We also use a continuous measure in the range $[0,1]$ that we call \textbf{Proportional} match $h_n$, i.e., the portion of LLM-generated distractors that match human-authored ones, defined as %\mh{I would name it $h_p$ but that's already used for partial...maybe we use superscripts here too?}
\begin{align*}
    h_n(D, \hat{D}) = \frac{\sum_{i=1} \mathbf{1}_{i: \hat{d}_i = d_i}}{N}
\end{align*}
where $\mathbf{1}$ denotes an indicator function. We report all metrics by averaging across all MCQs in the test set and scale the values of metrics by a factor of 100 into percentages. 

Additionally, we experiment with a non-standard, \textit{distribution-based} metric, which tries to predict how often a distractor is selected by real students. This metric is motivated by the observation (See Section~\ref{sec:qual} for a detailed qualitative analysis) that human-authored distractors are sometimes not plausible or complete: for some MCQs, there may only be one highly common error or misconception among real students, so teachers often have to come up with a few placeholders that will be selected by almost no one, while for other MCQs, there may be numerous plausible distractors that cannot all be included. Therefore, our goal is to use the percentages of students who selected each option to train a model that predicts how feasible a distractor is. Since we cannot reach high predictive accuracy on the real dataset we have, we relegate the details on this metric and experimental results to the Supplementary Material Section~\ref{app:pairrank}.

Furthermore, we experiment with a metric that evaluates the quality of LLM-generated distractors indirectly: we ask \texttt{GPT-4} to answer MCQs in the test set under two scenarios: one using LLM-generated distractors and the other using human-authored ones. We then calculate and compare the solve rates. We found that LLM-generated distractors using kNN, the best-performing method under alignment-based metrics, are more difficult (71\% solve rate) than the human-authored ones (72.5\%). This result implies that the LLM-generated distractors are not placeholders, which would make it very easy for \texttt{GPT-4} to select the key. However, solve rate cannot be used to evaluate the real quality of distractors and whether they reflect common errors or misconceptions among real students.

\subsection{Experimental Setup}
For all approaches except SB, we use a uniform format to represent the target MCQ. This format comprises a concatenation of 3 elements: the question stem, key, and its explanation. We use this structure since it encapsulates the most comprehensive information about the target MCQ. Furthermore, based on CoT, we instruct the LLM to first generate feedback message and then the distractor, which intends to simulate a reasoning pathway, providing a scaffold that guides the subsequent generation of plausible distractors. 
We use greedy decoding and a maximum output length of 350 tokens for distractor generation. Additional hyperparameters and model details are in Supplementary Material Section \ref{app:aes}. We also provide our prompts for CoT, RB, and kNN in Tables ~\ref{tab:CoT}, ~\ref{tab: RB}, and ~\ref{tab:kNN} respectively. 
% \ml{REMEMBER: use bold font ONLY when you define these approaches, not over and over again!}

\subsection{Results and Discussion}
\begin{table}
% \vspace{-0.2cm}
\centering
\scalebox{1.00}{
\begin{tabular}{cccc}
\toprule
\textbf{Approach}  & Exact   & Partial   & Proportional  \\
\midrule
$\text{kNN}$  & \textbf{10.95}   & \textbf{73.85}     & \textbf{38.52}   \\
$\text{CoT}$   & 4.24    & 65.02    & 30.39    \\
$\text{RB}$     & 4.95    & 57.95      & 27.21      \\
$\text{FT}$ & 2.83    & 57.95     & 25.32    \\
$\text{SB}$    & 0.00    & 10.25     & 3.65 \\ 
\bottomrule
\end{tabular}
}
\caption{Results on distractor generation on alignment-based metrics, where in-context learning with kNN example selection outperforms other approaches.}
\label{tab:methodology_comparison}
\end{table}

Table~\ref{tab:methodology_comparison} shows the results on distractor generation for the 5 approaches we explore. Overall, kNN outperforms the other approaches. This result is not surprising since examples that are textually similar to the target MCQ often contain distractors that correspond to plausible errors or misconceptions among real students for both MCQs. Therefore, the LLM can generate distractors that match the human-authored ones by simply replicating the style of the in-context examples. This approach is especially effective for MCQs that have highly similar structures and differ in only numerical values. 
%Other approaches score lower on these metrics, as the approaches do not find or have informative in-context samples. 

The advantage of CoT over FT reflects the strong mathematical reasoning capability of \texttt{GPT-4}, which results in a performance gap that not even fine-tuning \texttt{ChatGPT} on human-authored distractors can make up. Instead of acting as an oracle, RB underperforms this expectation and does not even outperform CoT, despite requiring significant human expertise and effort. This result is likely due to the fact that despite extensive effort in labeling error explanations, we cannot come up with a comprehensive list of them; as a result, many target MCQs are not matched with error explanations for \texttt{GPT-4} to select from. Overall, we observe that \texttt{GPT-4} can often generate mathematically valid distractors but is unaware of what errors or misconceptions are common among real students. Therefore, CoT and RB do not perform as well as kNN. Among all the approaches we explore, SB has by far the worst performance. This result is not surprising since when we train LLMs to answer math MCQs correctly, the generated incorrect answers are either only marginally different than the key or completely unrelated to the question stem. Therefore, the distractors generated by this approach lack coherent reasoning and fail to capture common errors or misconceptions among real students. 

\subsubsection{Ablation Study}
\begin{table}
% \vspace{0.2cm}
\centering
\scalebox{0.99}{
\begin{tabular}{lccc}
\toprule
{\textbf{Approach}}  & Exact   & Partial   & Proportional  \\
\midrule
$\text{kNN}^{\text{all}}$  & \textbf{10.95}   & \textbf{73.85}     & \textbf{38.52} \\
$\text{kNN}^{\text{key}}$  & 11.66    & 69.96     & 37.57  \\
$\text{kNN}^{\text{none}}$ & 9.54    & 67.84     & 35.81 \\
Random     & 2.12    & 54.77     & 23.56 \\
$\text{Prompt}^{\text{key}}$  & 8.48    & 66.43     & 34.04  \\
$\text{Prompt}^{\text{none}}$ & 3.89    & 44.52    & 21.20 \\
$\text{kNN}^{\text{all}}_{\neg T}$    & 3.18    & 58.30     & 26.38   \\
\midrule
$\text{FT}^{\text{gpt3.5}}$    & \textbf{2.83}    & \textbf{57.95}     & \textbf{25.32}   \\
$\text{FT}^{\text{mistral}}$    & 1.77    & 52.30     & 22.50   \\
\midrule
$\text{RB}^{\text{select}}$    & \textbf{4.95}    & \textbf{57.95}      & \textbf{27.21}    \\
$\text{RB}^{\text{random}}$    & 1.06    & 53.00     & 23.20   \\
\bottomrule
\end{tabular}
}
\caption{Results on ablation study on alignment-based metrics with different settings of kNN, FT, and RB.}
\label{tab:ablation}
\end{table}

In this ablation study, we investigate the impact of different configurations of kNN on its performance and summarize these results in the first part of Table~\ref{tab:ablation}. 
We explore how different ways of using different parts of the MCQ in the textual encoder for nearest neighbor search could affect kNN's performance. We experiment with 3 different settings: using just the question stem (\textbf{$\text{kNN}^{\text{none}}$}); using the question stem and key (\textbf{$\text{kNN}^{\text{key}}$}); and using the question stem, key, and its explanation (\textbf{$\text{kNN}^{\text{all}}$}), which is the best performing setting. For comparison, we also experiment with a simple random heuristic (\textbf{Random}) that chooses examples from the training set randomly without any specific criteria. We see that although using only the question stem captures the math skill covered by an MCQ and helps kNN find examples that have the same format as the target MCQ, adding the key and explanation helps kNN find better examples that use similar problem-solving strategies to the target MCQ. We also explore how different prompt formats could affect kNN's performance. We experiment with 3 different prompt formats. The best-performing setting ($\text{kNN}^{\text{all}}$) includes the question stem, key, and explanation for both the target MCQ and the in-context examples. The in-context examples also contain feedback on the distractors, and we ask the LLM to generate the feedback, followed by the distractor. The other settings are not to include feedback messages for the distractors and the explanation for the key (\textbf{$\text{Prompt}^{\text{key}}$}), and not including the key either (\textbf{$\text{Prompt}^{\text{none}}$}). We see that including the key significantly improves kNN's performance and asking the LLM to generate feedback followed by the distractor further improves performance. This result again reinforces the importance of math problem-solving strategies and CoT reasoning on the distractor generation performance. We also explore the impact of not allowing MCQs with the same topic to be selected as examples on kNN's performance (\textbf{$\text{kNN}^{\text{all}}_{\neg T}$}). We see that doing so results in a huge performance drop-off from $\text{kNN}^{\text{all}}$. This result suggests that most errors or misconceptions behind distractors are topic-specific and do not generalize across topics.

% for the key, which we dub \textbf{$\text{Prompt}^{\text{key}}$}, and not including the key either, which we dub \textbf{$\text{Prompt}^{\text{none}}$}. We see that including the key significantly improves kNN's performance and asking the LLM to generate feedback followed by the distractor improves performance. This result again reinforces the importance of math problem-solving strategies on distractor generation performance and chain-of-thought reasoning. We also explore the impact of not allowing MCQs with the same topic to be selected as examples of kNN's performance, which we dub $\text{kNN}^{\text{all}}_{\neg T}$. We see that doing so results in a huge performance drop-off from $\text{kNN}^{\text{all}}$. This result suggests that most mathematical errors behind distractors are topic-specific and do not generalize across topics. 

Next, we investigate the impact of different base LLMs on FT's performance and summarize these results in the second part of Table~\ref{tab:ablation}. We compare \texttt{ChatGPT} against \texttt{Mistral-7B} \cite{mistral}, which is one of the biggest open-sourced generative LLMs (\textbf{$\text{FT}^{\text{mistral}}$}). 
We see that \texttt{ChatGPT} outperforms \texttt{Mistral-7B} on all 3 alignment-based metrics. This result suggests that larger models that are better at mathematical reasoning are more likely to generate plausible distractors. We also investigate the impact of different error selection approaches on RB's performance and summarize these results in the third part of Table~\ref{tab:ablation}. We experiment with a variant of RB that randomly selects error explanations under the same math topic (\textbf{$\text{RB}^{\text{random}}$}) instead of asking \texttt{GPT-4} to select 3 relevant ones (\textbf{$\text{RB}^{\text{select}}$}). We see that asking the LLM to select error explanations outperforms selecting error explanations randomly, but not by a significant margin compared to other ablations. This result suggests that even though LLMs can generate many mathematically valid distractors, their ability to recognize which error explanations are popular among students is limited. 

\begin{table}[t!]
% \vspace{0.2cm}
\scalebox{0.85}{
\begin{tabular}{lcccc}
\toprule
\multicolumn{1}{c}{LLM}                         & Approach & Exact         & Partial        & Proportional   \\ \midrule
                            & $\textbf{kNN}$      & \textbf{8.83}          & \textbf{71.73}          & \textbf{38.52}          \\
\multicolumn{1}{c}{GPT-4}   & $\text{CoT}$      & 4.24 & 65.02 & 30.39 \\
                            & $\text{RB}$       & 4.95          & 57.95          & 27.21          \\ \midrule
                            & $\textbf{kNN}$      & \textbf{10.95} & \textbf{73.85} & \textbf{38.52} \\
\multicolumn{1}{c}{ChatGPT} & $\text{CoT}$      & 1.06          & 48.41          & 21.67          \\
                            & $\text{RB}$       & 1.77          & 50.53          & 23.09          \\
                            & $\text{FT}$       & 2.83          & 57.95          & 25.32          \\ \midrule
                            & $\text{kNN}$     &1.77           &31.10           &13.90            \\
\multicolumn{1}{c}{Mistral}  & $\text{CoT}$      &0.0          &8.83            &3.65           \\
                            & $\text{RB}$       &0.0          &18.37         &7.07           \\
                            & $\textbf{FT}$       & \textbf{1.77} & \textbf{52.30} & \textbf{22.50} \\ \bottomrule
\end{tabular}}
\caption{Results on kNN, CoT, RB, and FT on alignment-based metrics with different base LLMs: \texttt{GPT-4}, \texttt{ChatGPT}, and \texttt{Mistral}.}
\label{tab:ablation2}
\end{table}

Furthermore, we investigate the impact of different base LLMs on all approaches' performance except SB and summarize these results in Table~\ref{tab:ablation2}. We compare 3 base LLMs: \texttt{GPT-4} \footnote{As of now, Openai does not allow fine-tune \texttt{GPT-4}.}, \texttt{ChatGPT}, and \texttt{Mistral}. We see that kNN outperforms CoT and RB across all base LLMs. This result suggests that kNN is a promising approach since in-context examples provide valuable information to the LLM on the nature and format of the distractor generation task. We see that \texttt{GPT-4} significantly outperforms \texttt{ChatGPT}, which significantly outperforms \texttt{Mistral} on CoT and RB. This result suggests that, among the 3 base LLMs evaluated, \texttt{GPT-4} possesses the most robust mathematical reasoning capability, followed by \texttt{ChatGPT}, which possesses a better mathematical reasoning capability than \texttt{Mistral}. We see that FT achieves better performance than kNN with \texttt{Mistral}. This result suggests that the pre-trained \texttt{Mistral} initially lacks mathematical reasoning capability, and the fine-tuning process significantly enhances its mathematical reasoning capability.

\subsection{Qualitative Analysis}
\label{sec:qual} 
\begin{table} [t!]
% \vspace{0.2cm}
    \begin{tabular}{p{7cm}}
        \toprule
         {\textbf{Target}} \\
         \small Quesiton stem: which multiplier can be used to find the value after an amount has decreased in value by 8\% for 4 years? \\
         \small Explanation: As its is a decrease, we need 100\% - 8\% which is 92\% which is the same as 0.92. We then use the number of years as the power of 4. \\
         \small Answer: $\times 0.92^{4}$ \\
        \midrule
         {\textbf{Example 1}} \\ 
         \small Quesiton stem: which multiplier can be used to find the value after an amount has decreased in value by 5\% for 5 years? \\
         \small Explanation: As its is a decrease, we need 100\% - 5\% which is 95\% which is the same as 0.95. We then use the number of years as the power of 5. \\
         \small Answer: $\times 0.95^{5}$ \\
        \midrule
         {\textbf{Example 2}} \\ 
         \small Quesiton stem: the value of a laptop that initially cost \$1100, declines in value by 15\% a year. if you wanted to calculate the value of the tablet at the end of 6 years, what number would replace the square? $1100 \times \square^{6}$ \\
         \small Explanation: As the value decreases by 15\%, we have 100\% - 15\% = 85\% = 0.85 as the multiplier. \\
        \small Answer: 0.85 \\
        \midrule
         {\textbf{Example 3}} \\ 
         \small Quesiton stem: a car depreciates in value by 15\% each year. if a car was bought for \$3500, which of the following calculations would find the new value of the car after 3 years?\\
         \small Explanation: The multiplier is 1 - 0.15 = 0.85, and as we are using compound interest, we raise this to the power of 3. \\
        \small Answer: $3500 \times 0.85^{3}$ \\
        \bottomrule
    \end{tabular}
    \captionof{table}{Three in-context learning examples retrieved by $\text{kNN}$; we see that Example 1 is very similar to the target MCQ, except for different numerical values.}
    \label{table:distractor-kNN-SEKFD}
\vspace{-0.1cm}
\end{table}

We now qualitatively investigate the distractors generated by the best approach, $\text{kNN}$, to extract some insights on the distractor generation task and how to improve performance. We group the 283 total MCQs in the test set into 4 categories, according to the number of LLM-generated distractors that match the human-authored ones, from 0 to 3.

For the group where all LLM-generated distractors match the human-authored ones (3 out of 3), we find that, in all but 2 of the 28 such cases, there is an in-context example that is very similar to the target MCQ, with the only difference being different numerical values or named entities. See Table~\ref{table:distractor-kNN-SEKFD} for an example. However, this situation sometimes appears in other groups too, which is perhaps surprising since it implies that the presence of a near-identical in-context example alone is not sufficient for an LLM to generate plausible distractors. We investigate further into such cases and find that even for two MCQs with near-identical question stem, their sets of distractors and the errors or misconceptions underlying each distractor may differ even though both are plausible. This situation occurs when there are more than 3 plausible errors or misconceptions given a question stem. 
\begin{table} [h]
    \centering
    \begin{tabular}{p{7.2cm}}
    \toprule 
    \textbf{Question Stem} \\
    % \midrule
    Craig and Isaac share some fruit. Isaac gets three-quarters of the fruit. In what ratio do they share the fruit? (Isaac's part second) \\
    \midrule
    \textbf{Key}\\  1 : 3\\
    \midrule
    \textbf{LLM-generated Distractors} \\ $3:1$ $\quad$ $3:4$ $\quad$ $4:1$ $\quad$\\
    \midrule
    \textbf{Human-authored Distractors} \\ $1:4$ $\quad$ $1:2$ $\quad$ $4:3$ $\quad$\\
    \bottomrule
    \end{tabular}
    \caption{Example of LLM-generated distractors that are mathematically valid and plausible but do not match human-authored ones.}
    \label{table:distractor-example-multiples}
    % \vspace{-0.1cm}
\end{table}

For the group where none of the LLM-generated distractors match the human-authored ones, we randomly select 20 of the 78 cases to analyze. We find that in 14 of the 20 cases (70\%), the LLM-generated distractors are plausible, and the human-authored ones are not superior to the LLM-generated distractors. See Table~\ref{table:distractor-example-multiples} for an example. While this observation is entirely subjective, it highlights that alignment-based metrics may not be an appropriate metric to measure the quality of LLM-generated distractors because human-authored ones may not be naturally optimal. This observation is also part of our motivation in developing \textit{distribution-based} metrics to predict how likely a LLM-generated distractor will be selected by real students with insufficient knowledge. Moreover, since many LLM-generated distractors are plausible even if they are not the same as the human-authored ones, there is promise in using automated distractor generation for teacher support during the generation of MCQs.
\begin{table} [h!]
    \centering
    \begin{tabular}{p{7.2cm}}
    \toprule
    \textbf{Question Stem} \\
    % \midrule
    Convert $0.6$ to a fraction in its simplest form. \\
    \midrule
    \textbf{Key}\\ $\frac{3}{5}$\\
    \midrule
    \textbf{LLM-generated Distractors}\\ $\frac{6}{10}$ $\quad$ $\frac{5}{3}$ $\quad$ $\frac{6}{5}$ $\quad$\\
    \midrule
    \textbf{Human-authored Distractors}\\ $\frac{6}{10}$ $\quad$ $\frac{60}{100}$ $\quad$ $\frac{1}{6}$ $\quad$\\
    \bottomrule
    \end{tabular}
    \caption{Example of LLM-generated distractors where the plausible one, $\frac{6}{10}$ matches the human-authored ones, while the rest of human-authored ons are placeholders. In this case, $\frac{6}{10}$ is selected by 28\% of students while other distractors are rarely being selected.}
    \label{table:distractor-example-fraction-simplify}
% \vspace{-0.5cm}
\end{table}

Finally, for the group where 1 or 2 LLM-generated distractors match the human-authored ones, we examine which human-authored distractor(s) are generated. We find that in many cases, the human-authored distractors that match the LLM-generated ones seem to reflect common errors or misconceptions among real students, while the others do not. See Table~\ref{table:distractor-example-fraction-simplify} for an example. This observation is further supported by selections made by real students, where the distractors that correspond to the common errors or misconceptions are selected by more students in 44 of 108 (40.7\%) cases and 46 of 63 (73\%) cases for 1 and 2 matches, respectively, while the rest are rarely being selected. This result suggests that many MCQs have 1 or 2 highly plausible distractors while the others are placeholders. Again, using human-authored ones as the ground truth on alignment-based metrics is not ideal, which justifies our motivation in developing the distribution-based metric. 

\subsection{Human Evaluation}
We conduct a human evaluation to assess the quality of LLM-generated distractors. This evaluation is motivated by observations from the qualitative analysis that the generated distractors are often plausible even though they may be different from human-authored ones.

\subsubsection{Evaluation Design}
We recruit 2 graduate students who have experience teaching math or related topics as human evaluators. They are presented with the same set of 20 MCQs that are randomly sampled from the test set, each accompanied by a mixture of 4 or 6 distractors. To ensure a balanced assessment, half of these are LLM-generated distractors, while the remaining are human-authored ones. To eliminate any potential ordering bias, the sequence of the distractors is randomized for each question. They are asked to rate the distractors on two aspects: mathematical validity (\textbf{validity}) and plausibility for middle school math students (\textbf{plausibility}). Validity measures the degree of a distractor that is relevant to the question stem and can be tangibly reached by some incorrect reasoning. Plausibility measures how likely a distractor is to be selected by real students. Each aspect is scored on a scale from 1 to 5, with 1 being the lowest: a distractor that is irrelevant to the question stem or one that no real students would select, while 5 being the highest: a distractor that is highly relevant to the question stem or one that is highly likely to trick real students with insufficient math knowledge into selecting it. Additional evaluation setup details are in Supplementary Material Section \ref{app:hed}.

\subsubsection{Evaluation Result and Discussion}
\begin{table}[t!]
% \vspace{0.2cm}
\centering
\scalebox{0.93}{
\begin{tabular}{lcccc}
\toprule
& \multicolumn{2}{c}{\textbf{QWK}} & \multicolumn{2}{l}{\textbf{Average Ratings}} \\
        \cmidrule(lr){2-3} \cmidrule(lr){4-5}
        & LLM & Human & LLM & Human \\
\midrule
\multicolumn{1}{c}{Validity}     & 0.34        & 0.23          &  3.28              & 3.99$^*$                 \\
\multicolumn{1}{c}{Plausibility} & 0.54        & 0.54          & 2.68               & 3.72$^*$ \\
\bottomrule
\end{tabular}
}
\caption{QWK and average ratings among human evaluators on LLM-generated and human-authored distractors for validity and plausibility. Under a Student's t-test, human evaluators prefer human-authored distractors with statistical significance ($p<0.05^*$).}
\label{tab:human_eval}
\end{table}

Table~\ref{tab:human_eval} shows the inter-rater agreement, measured using quadratic weighted Kappa (QWK) \cite{qwk} and the average rating across 2 human evaluators for both LLM-generated and human-authored distractors. The QWK scores indicate a fair to moderate level of agreement between two human evaluators regarding both the validity and plausibility aspects of distractors. This observation suggests that measuring the quality of distractors based on their validity and plausibility is consistent at certain level and can be used in future assessments of distractors. We conduct a Student's t-test \cite{ttest} to compare the ratings for LLM-generated and human-authored distractors and find that in both aspects, there is a statistically significant difference ($p < 0.05$). This result shows that human evaluators think human-authored distractors are better than LLM-generated distractors in both aspects. Furthermore, we observe that the gap between LLM-generated distractors and human-authored ones is much bigger for plausibility than validity. This observation indicates that LLMs exhibit a higher proficiency in generating mathematically valid distractors compared to anticipating common errors or misconceptions among real students. See Table~\ref{table:valid} for an example. This result is not surprising since LLMs, which have not been extensively trained on erroneous answers provided by real students, may struggle to anticipate the various ways in which students are prone to making errors or student misconceptions. Therefore, there is still considerable room for improvement for LLMs in their capacity to anticipate errors or misconceptions among real students.

\begin{table} [t]
% \vspace{0.2cm}
    \centering
    \begin{tabular}{p{7.1cm}}
    \toprule
    \textbf{Question Stem} \\
    % \midrule
    Solve this problem: $\sqrt[3]{216} \: = \: ?$\\
    \midrule
    \textbf{LLM-generated feedback message}\\ I think you have multiplied by 3. The question is asking for the cube root. \\
    \midrule
    \textbf{LLM-generated distractor}\\ 648 \\
    \midrule
    \textbf{LLM-generated feedback message}\\ I think you have written the digits as a new number. The question is asking for the cube root. \\
    \midrule
    \textbf{LLM-generated distractor}\\ 2163 \\    
    \bottomrule
    \end{tabular}
    \caption{Examples of LLM-generated distractors, which, from a purely mathematical perspective, seem valid as the errors suggest misunderstandings of the cube root operation as either multiplying by 3 or appending a 3 to the original number. However, these two distractors do not effectively reflect the common errors or misconceptions among real students.}
    \label{table:valid}
\end{table}

% without thoroughly trained on erroneous answers from by real students, would not be able to adapt well to ways in which students would make errors\jk{wow this sentence is so poor. would .... would ... Use ChatGPT for this situation.}. Therefore, there is still considerable room for improvement for LLMs in their capacity to understand student errors in MCQ. 
%We conduct a paired t-test \cite{pairt} to compare the average ratings for Validity and Plausibility on LLM-generated distractors, and find that there exists statiscally significant difference ($p < 0.05$). This result shows that while the LLM-generated distractors are valid for the question stem, they do not necessarily reflect the common errors among real students.

% Moreover, we conduct a Student's paired t-test \cite{ttest} to compare the average ratings for Validity and Plausibility on LLm-generated  and find that on both sets of distractors, there is statistical significant difference ($p < 0.05$). This result shows that the human evaluators consider the 

\section{Conclusions and Future Work}
In this paper, we explore automated distractor generation for math multiple-choice questions via large language models. We conduct experiments on a real-world math MCQ dataset and find that the in-context learning-based approach kNN, achieves the best performance when compared to other approaches such as fine-tuning, chain-of-thought prompting, and various baselines. We also conduct human evaluation and observe that LLMs are capable of generating mathematically valid distractors but are not fully aware of common errors or misconceptions among real students. 
Our initial exploration of this task opens up many avenues for future work. For example, we need to further refine the distribution-based metrics that predict the percentage of students who select each distractor. We also need to develop modified text encoding approaches that are closely aligned with errors or misconceptions among real students for in-context example selection. Furthermore, we aim to explore the generation of distractors, each of which corresponds to a specific error or misconception, in a planned and controllable way \cite{wang2021math,zhang2023interpretable}, possibly taking student knowledge state into account \cite{liu2022open}, as well as the generation of high-quality feedback messages for each distractor \cite{scarlatos2024improving}. Finally, extending our work from multiple-choice questions to open-ended questions is important, since open-ended student responses contain much more detailed information on their errors \cite{zhang2021math,zhang2022automatic,mcnichols2023algebra}. 

\section{Acknowledgement}
The authors thank Schmidt Futures and the NSF (under grants IIS-2118706 and IIS-2237676) for supporting this work.

% Moreover, we want to explore two potential challenging tasks: 1) the distractor generation task can be treated as a controlled generation task \cite{cg} because we observe that there exists an inherent error for each question stem and distractor pair. We aim to explore effective approaches to learn the representation of these inherent errors and generating the distractor that align with the given error; 2) generating suitable feedback messages for distractors \cite{lan-grading-and-feedback, zhang2021math} because providing feedback to students who select distractors can help them identify their errors, which may expedites their learning process.

\clearpage
\newpage

\section*{Limitations}
Being the attempt at the task of generating plausible distractors for math MCQs using LLMs,  we find several limitations in our current setup. First, we find that some human-authored distractors are merely placeholders that do not reflect common errors or misconceptions among real students, and using them as in-context demonstrations may lead to the generated distractors also not reflecting common errors or misconceptions among real students. Second, the \textit{alignment-based} metrics may not accurately measure the quality of LLM-generated distractors because some MCQs may have only 1 or 2 plausible distractors and some MCQs have more than 3 plausible distractors. Third, we acknowledge that our human evaluation sample size is small, and should ideally be increased for future studies in order to receive more accurate results.

\section*{Ethical Considerations}
The focus of our work is to automatically generate plausible distractors for math MCQs using LLM. By automating part of the MCQ generation, we aim to save educators and teachers from time-consuming MCQ generation and allow them to dedicate more effort to teaching and student engagement. Based on our analysis on the generated distractors, we acknowledge that not every distractor generated by our work is plausible. Therefore, we strongly advise that our work should be adopted as an auxiliary tool in the generation of MCQs. All automatically generated distractors should undergo a careful review by educators and teachers before being utilized in real tests for students.
\bibliography{anthology, custom}


%\clearpage

\newpage 

\noindent {\Large \textbf{Supplementary Material}}

\appendix

\section{Distribution ranking metric}
\label{app:pairrank}

Since our qualitative analysis in Section~\ref{sec:qual} found that human-authored distractors are sometimes unplausible or incomplete, using them as the ground truth is not ideal. Therefore, we explore a distribution-based metric to evaluate the quality of LLM-generated distractors, based on one intuition: good distractors are ones that are likely going to be selected by many real students. Therefore, our goal is to train a model that can predict the portion of students that select each option in an MCQ. However, due to the highly noisy nature of this distribution, we opt to train a model that predicts the more often selected distractor among a pair, given a question stem, which is similar to the pairwise preference reward model in reinforcement learning from human feedback (RLHF) \cite{christiano2017deep}. After training such a model, we can use it to compare generated distractors to human-authored ones in head-to-head matchups, giving us a proxy for how good an LLM is in terms of generating distractors that are likely to be selected by students.

Formally, we train an LLM-based model $\operatorname{r}_\phi(d_1, d_2, s, k, e_k) \rightarrow \{d_1, d_2\}$, where $\phi$ denotes the set of model parameters. We train this model by first constructing a dataset of all pairs of human-authored distractors for each MCQ and include both orders of each pair to avoid ordering bias, resulting in $N \times {3 \choose 2} \times 2$ total pairs, where $N$ denotes the number of MCQs. Each pair is associated with a binary-valued label indicating whether $d_1$ or $d_2$ is selected by more students, which we can calculate from the student response records in our dataset. We then use this dataset to fine-tune an LLM in a text generation task, where the LLM receives the question and distractor information in its prompt and outputs its preference. We show our prompt for this task in Table \ref{tab:prompt-ranking}. 

We use the same train/test split as the distractor generation experiments, and reserve 20\% of the train split for validation after each epoch and early stopping. We fine-tune the \texttt{mistralai/Mistral-7B-v0.1} model, which contains 7 billion parameters, from HuggingFace \cite{huggingface2019} using LoRA \cite{lora} with adaptors on the \texttt{q\_proj}, \texttt{k\_proj}, \texttt{v\_proj}, and \texttt{o\_proj} matrices, set $r=32$, $\alpha=16$, $\text{dropout}=0.05$, and use 8-bit quantization. We train the model using the AdamW optimizer for 10 epochs with a learning rate of 3e-5, a batch size of 16, accumulate gradients for 4 batches. The model converges on the validation set after 6 epochs. The GPU we use to train the model is \texttt{NVIDIA RTX A6000}. The training process is completed in 10 hours. When evaluated on the test set, the ranking model correctly identifies the preferred distractor $61.60\%$ of the time (random guessing corresponds to $50\%$ accuracy). This accuracy is low overall but high on subsets of distractor pairs whose student selection percentages differ by a large margin: on pairs with a larger than $20\%$ margin, which accounts for $6\%$ of pairs, the accuracy jumps to $74.47\%$. This result is not surprising since the selection percentage data is very noisy. 

%We also evaluate on subsets of the test set where the difference between the portion of times distractors are chosen is greater than some cutoff. When using a cutoff of $5\%$, comprising $51\%$ of the pairs in the test set, the model achieves $66.36\%$ accuracy. Increasing the cutoff to $15\%$ ($12\%$ of pairs) brings accuracy to $71.08\%$, and further increasing the cutoff to $20\%$ ($6\%$ of pairs) brings accuracy to $79.79\%$. These results show that the ranking model becomes more accurate when one distractor is strongly preferred over the other. This trend can partly be explained by noise in the data; when distractors are similarly likely to be chosen by students, it is possible that either could be chosen more frequently in the data. It is also likely that the model learns to identify common cases of certain distractors being consistently preferred over others. \al{do we need a qualitative analysis, like identifying common fail cases or which questions do better than others?}

Using this trained model, we can evaluate the quality of LLM-generated distractors: we compare all possible head-to-head matchups between generated distractors and human-authored ones, and record the portion of times that the generated distractors are preferred by the ranking model. If the two distractors are the same then we record a tie. In cases where the generated distractors are invalid or repeated, we treat them as \texttt{null} and record a win for the human-authored ones. Formally, we define a preference score as
%We additionally handle cases where there are not 3 valid and unique generated distractors for a question; we remove any generated distractors that are the key for the question, remove duplicate generated distractors, and consider 0 distractors to be generated if the generative model's output does not match the expected template. We pad all resulting sets of less than 3 generated distractors with \texttt{null} values, and consider the human-written distractor to be preferred when compared to a \texttt{null} distractor. We formally define a final system-level score as follows:
\begin{align*}
    & s = \frac{1}{18N}\sum_{i=1}^N \sum_{a=1}^3 \sum_{b=1}^3 \operatorname{r}^{(i)}_\phi(\hat{d}_a^{(i)}, d_b^{(i)})\\
    & \quad\quad\quad\quad\quad + (1 - \operatorname{r}^{(i)}_\phi(d_b^{(i)}, \hat{d}_a^{(i)})),\\
    & \operatorname{r}^{(i)}_\phi(d_1, d_2) = \begin{cases} 0.5 & d_1 = d_2 \\ 1 & d_2 \text{ is \texttt{null}} \\ 0 & d_1 \text{ is \texttt{null}} \\ p & \text{otherwise} \end{cases},\\
    & p = \mathbf{1}_{\operatorname{r}_\phi(d_1, d_2, s^{(i)}, k^{(i)}, e_k^{(i)}) = d_1},
\end{align*}
where $\hat{d}$ are generated distractors. This score has a range of $[0,1]$ where higher values indicate LLM-generated distractors are likely to be selected by more students than the human-authored ones. We found that kNN scores $0.46$ on the test set, which indicates that the distractors it generates are almost as plausible to students as human-authored ones. 

We emphasize that this evaluation metric should only be considered exploratory due to several obvious limitations. First, student option selection percentages create noisy labels for the ranking model, limiting its accuracy. Second, using the overall selection percentages also ignores the individual learning context of each student since students with different knowledge levels may have different tendencies among MCQ options. Therefore, we leave a more thorough treatment of the distribution-based metric to future work. 

\section{Hyperparameters and Implementation Details}
\label{app:aes}
For fine-tune (FT) approach, We fine-tune 2 large language models. We fine-tune the \texttt{mistralai/Mistral-7B-Instruct-v0.2} model, which is the latest \texttt{Mistral} model and contains 7 billion parameters, from HuggingFace using LoRA with adaptors on the \texttt{q\_proj}, \texttt{k\_proj}, \texttt{v\_proj}, and \texttt{o\_proj} matrices, set $r=32$, $\alpha=16$, $\text{dropout}=0.05$, and use 8-bit quantization. We use 20\% of the training set for validation. We train the model using the AdamW optimizer with $\text{weight decay}=0.0$ and $\text{gradient clip}=1.0$ for 8 epochs with a learning rate of 8e-5, a batch size of 16, and accumulated gradients for 4 batches. The selection of the aforementioned hyperparameters is guided by exploratory evaluations and no substantial hyper-parameter search is conducted. The GPU we use to train the model is \texttt{NVIDIA RTX A6000}. The training process is completed in 2 hours and 55 minutes. We fine-tune the \texttt{ChatGPT} model using the first 200 data points from the training set. We train the model using the OpenAI's default fine-tuning settings, which we find to provide the best performance via OpenAI API. The training process is completed in 20 minutes. We use the \texttt{scikit-learn} \cite{sklearn} implementation to calculate QWK, and use the \texttt{scipy} \cite{scipy} implementation to calculate Student's t-test. For prompting \texttt{GPT-4} and \texttt{ChatGPT} using OpenAI API, we use $\text{temperature}=0$, $\text{max\_tokens}=350$, $\text{top\_p}=1$ as our setup for greedy decoding. All our experiments are implemented in Python or Pytorch code, and We note that all software employed in this work is open-source, or the license is unspecified. 

\section{Human Evaluation Details}
\label{app:hed}
In this work, we obtained approval from the ethics review board for human evaluation. We show the evaluation instructions to human evaluators in Table ~\ref{tab:ins}. We do not provide any compensation for human evaluators because their participation is entirely voluntary and we appreciate their contribution to this work.

\clearpage
\onecolumn
\section{Prompt Format}
We provide the prompts for CoT, RB, and kNN in the work below. We use $<>$ to indicate that a variable is filled in dynamically.
\begin{longtable}{p{4.5cm}p{10.5cm}}
    \toprule
    \textbf{Prompt} & You are given the following math question along with the correct answer and explanation. Please use the following template to give 3 alternative incorrect answers to be used as multiple-choice options in a multiple-choice exam. Prior to the incorrect answer, provide feedback to be displayed to the student as an explanation of why that is not the correct answer.\newline 
    [Template]\newline 
    Distractor1 Feedback:\newline
    Distractor1:\newline
    Distractor2 Feedback:\newline
    Distractor2:\newline
    Distractor3 Feedback:\newline
    Distractor3:\newline
    Question: <question>\newline
    Explanation: <explanation>\newline
    Answer: <answer>\\
    \bottomrule
    \caption{CoT prompt format}
    \label{tab:CoT} \\
\end{longtable}

\begin{longtable}{p{4.5cm}p{10.5cm}}
    \toprule
    \textbf{Prompt} & You are given the following math question along with the correct answer, explanation, and a list of errors. Please follow the template to first select 3 most likely errors for this question and use the selected errors to generate 3 alternative incorrect answers to be used as multiple-choice options in a multiple-choice exam. Prior to the incorrect answer, provide feedback to be displayed to the student as an explanation of why that is not the correct answer. If the list of errors is not given, generate 3 errors instead and do not contain any explanation in the 3 incorrect answer.\newline 
    [Template]\newline 
    Error1: \newline
    Error2: \newline
    Error3: \newline
    Distractor1 Feedback:\newline
    Distractor1:\newline
    Distractor2 Feedback:\newline
    Distractor2:\newline
    Distractor3 Feedback:\newline
    Distractor3:\newline
    Question: <question>\newline
    Explanation: <explanation>\newline
    Answer: <answer>\newline
    Error list: <error list>\\
    \bottomrule
    \caption{RB prompt format}
    \label{tab: RB} \\
\end{longtable}

\begin{longtable}{p{4.5cm}p{10.5cm}}
    \toprule
    \textbf{Prompt} & Question: <in-context question>\newline 
    Explanation: <in-context explanation>\newline
    Answer: <in-context answer>\newline
    Distractor1 Feedback: <in-context distractor1 feedback>\newline
    Distractor1:<in-context distractor1>\newline
    Distractor2 Feedback: <in-context distractor2 feedback>\newline
    Distractor2:<in-context distractor2>\newline
    Distractor3 Feedback:<in-context distractor3 feedback>\newline
    Distractor3:<in-context distractor3>\newline
    [stop]\newline
    Question: <target question>\newline
    Explanation: <target explanation>\newline
    Answer: <target answer>\\
    \bottomrule
    \caption{kNN prompt format, in practice, we use 3 in-context examples}
    \label{tab:kNN} \\
\end{longtable}

\onecolumn
\section{Ranking Metric Examples}
\begin{longtable}{p{4.5cm}p{10.5cm}}
    \toprule
    \textbf{Prompt} & A teacher assigns the following math multiple choice question to a class of middle school students.\newline

    Question: $\frac{3}{5} $ of 50 $= \frac{6}{10}$ of $\square$\newline
    Correct Answer: 50\newline
    Solution: 3/5 and 6/10 are equivalent, so 3/5 of 50 is the same as 6/10 of 50.\newline
    
    Here are 2 incorrect options that some students choose:\newline
    Option A: 30\newline
    Option B: 18\newline
    Which incorrect option are the students more likely to pick?  \\
    \midrule
    \textbf{Output} & Preferred Answer: A \\
    \bottomrule
    \caption{Example prompt and output for the ranking model used in the distribution ranking metric.}
    \label{tab:prompt-ranking} \\
\end{longtable}
% \begin{longtable}{p{4.5cm}p{10.5cm}}
%     \toprule
%     \textbf{Prompt} & A teacher assigns the following math multiple choice question to a class of middle school students.\newline

%     Question: $\frac{3}{5} $ of 50 $= \frac{6}{10}$ of $\square$\newline
%     Correct Answer: 50\newline
%     Solution: 3/5 and 6/10 are equivalent, so 3/5 of 50 is the same as 6/10 of 50.\newline
    
%     Here are two incorrect options that some students choose:\newline
%     Option A: 30\newline
%     Option B: 18\newline
%     Which incorrect option are the students more likely to pick?  \\
%     \midrule
%     \textbf{Output} & Preferred Answer: A \\
%     \bottomrule
%     \caption{Example prompt and output for the ranking model used in the distribution ranking metric.}
%     \label{tab:prompt-ranking} \\
% \end{longtable}

% \onecolumn
\section{Instruction}
\begin{longtable}{p{15.5cm}p{10.5cm}}
    \toprule
    You are given a csv file. Each row corresponds to a question stem and a distractor.\\
    Your job is to rate the distractor on two aspects: mathematical validity and plausibility for middle school math students. \\
    Mathematical validity measures whether a distractor is relevant to the question stem and can be tangibly reached by some incorrect reasoning. Mathematical validity is scored on a scale from 1 to 5, where 1 indicates a distractor that is irrelevant to the question stem, and 5 indicates a distractor that is highly relevant to the question stem.\\
    Plausibility measures how likely a distractor is to be selected by middle school students learning math. Plausibility is scored on a scale from 1 to 5, where 1 indicates that no student would select it and 5 indicates that the distractor is highly likely to trick students with insufficient math skills into selecting it.\\
    please use \texttt{numbers} on mac to rate distractors and give 1 and 1 for both metric if the distractor is the correct answer.\\
    Your ratings will be used to quantitatively measures and analyzes the quality of distractors on validity and plausibility.\\
    \caption{Instruction for Human Evaluation}
    \label{tab:ins}
\end{longtable}

\end{document}